\documentclass[acmtog,nonacm]{acmart}
\usepackage{hyperref}
\usepackage{url}
\usepackage{graphicx}
\usepackage{booktabs}   %
\usepackage{listings}   %
\usepackage{colortbl}   %
\usepackage{dblfloatfix}
\usepackage{afterpage}  %
\definecolor{bestcell}{RGB}{158,202,225}    %
\definecolor{secondcell}{RGB}{222,235,247}  %

\newcommand{\paper}[2]{#1~\citep{#2}}
\newcommand{\methodname}{K2V}

\newcommand{\FO}[1]{#1}

\begin{document}

\title{K2V: Generative Keyframing For Physically Plausible Video Generation}

\author{Felix O'Mahony}
\email{foo22@cam.ac.uk}
\orcid{TODO}
\correspondingauthor
\affiliation{%
  \institution{University of Cambridge}
  \city{Cambridge}
  \country{UK}
}

\author{Daniel Olmeda Reino}
\email{daniel.olmeda.reino@toyota-europe.com}
\orcid{TODO}
\affiliation{%
  \institution{Toyota Motor Europe}
  \city{Brussels}
  \country{Belgium}
}

\author{Roberto Cipolla}
\email{rc10001@cam.ac.uk}
\orcid{TODO}
\affiliation{%
  \institution{University of Cambridge}
  \city{Cambridge}
  \country{UK}
}

\author{Ayush Tewari}
\email{at2164@cam.ac.uk}
\orcid{TODO}
\affiliation{%
  \institution{University of Cambridge}
  \city{Cambridge}
  \country{UK}
}

\renewcommand{\shortauthors}{O'Mahony et al.}

\begin{abstract}
Video diffusion models are able to generate samples with high-quality appearance and textural qualities. 
However, content produced by these methods is known to exhibit significant physical errors~\citep{motamed_2026}. 
We propose to resolve these issues by redistributing the task of physical reasoning onto a language model-backed keyframe generation module\FO{, which allows for explicit reasoning in language, and fine-grained control over the appearance of the video at important moments in the video}. 
We then condition a video generation model with these keyframes, reducing the role of the diffusion model to the simpler task of interpolating between physically-correct intermediate states, rather than inferring an entire sequence with diffusion. 
Results demonstrate that our approach significantly outperforms existing approaches, achieving state-of-the-art performance on the \paper{Physics-IQ Verified benchmark}{physicsiq_verified_2026} and \paper{PhyGenBench}{phygenbench_2024}.
\end{abstract}

\begin{teaserfigure}
    \centering
    \includegraphics[width=\linewidth]{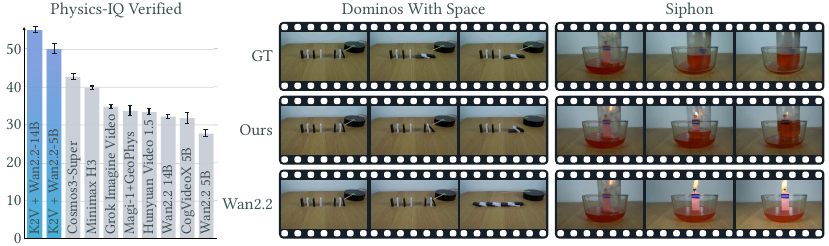}
    \caption{%
        Our approach generates keyframes to serve as stronger conditioning for a video model, leading to an approach which outperforms existing methods both qualitatively and quantitatively.
    }
    \label{fig:teaser}
\end{teaserfigure}

\maketitle

\section{Introduction}

Video generation models are increasingly positioned as `world simulators'~\citep{sora_2024}: models which, given a starting observation or a text prompt alone, can evolve a scene forward in time to give plausible future states.
The framing originates with \paper{Sora}{sora_2024}, which points to properties such as 3D consistency and object permanence as evidence that a generator trained only on pixels acquires a model of the world it depicts.
Other works build on this premise. \paper{Genie}{genie_2024} uses a video model directly as an interactive environment which a user acts in, and the model renders the consequences. \paper{Cosmos 3}{cosmos3_2026} positions its `omnimodal world model' as a potential backbone for autonomous physical agents.
However, video diffusion models are known to perform poorly when tasked with generating physically plausible videos. 
\paper{Physics-IQ}{motamed_2026} defines a world model as a video model which has internalised physical principles, and shows that existing models are severely limited in the satisfaction of this definition despite the high visual fidelity of their outputs.
The same conclusion is reached by \paper{VideoPhy-2}{videophy2_2025}, \paper{PhyGenBench}{phygenbench_2024} and \paper{PhyWorldBench}{phyworldbench_2025}, which together demonstrate failures across a broad spectrum of physical phenomena.

In this paper we introduce Keyframes to Video (\methodname{}), a novel approach to video generation which simplifies the task required of a video model to interpolation between keyframes, offloading physical inference to a separate keyframe generation module. 
Our first insight is that video models are more successful at reconstructing the physical motion connecting two explicit keyframes than at inferring plausible physical motion from a starting state entirely in diffusion space. 
We identify a simple training-free approach to condition an existing video generation model, \paper{Wan2.2}{wan_2025}, to perform the task of keyframe interpolation. 
Our second insight is that the physical reasoning required to plan those keyframes is already present in language models.
We therefore introduce a novel keyframe generation module, in which a language model plans the physical evolution of the scene.
This approach redistributes the load of physical planning away from a video diffusion model and onto a language model, which can then finely control an image edit model to produce the intended physical effect.

We argue that this division between planning keyframes and interpolating between them is the right allocation of effort for physically plausible generation, since the keyframe module can concentrate compute, both in tokens and in keyframe density, on the scenes and tasks which are most complex. 
A per-scene analysis of our results on the Physics-IQ Verified benchmark confirms that our approach does scale its effort with scene complexity in this way.

\methodname{} contributes to an existing line of research which enables physically plausible generation by delegating the physics to a separate model. 
Existing works, such as \paper{PhysGen}{physgen_2024} and \paper{VDAWorld}{vdaworld_2026}, leverage explicit physics simulators to propagate the physical properties of scenes forward in time, before rendering these scenes or using the output of these simulations to condition a video model.
\paper{NEWTON}{newton_2026} instead equips an agent with a sandboxed interpreter in which the relevant physical quantities are computed explicitly.
However, reliance on simulators carries significant limitations. 
Using simulators restricts the operating domain of these methods to settings where simulators are accurate and available, and the methods must first create a mapping from image to code, which can introduce errors. 
We dispense with simulators entirely, relying on the physical reasoning of language models in their place. 
In contrast to these works, we also reformulate the task posed to the video generator into one of interpolation, rigidly imposing the conditions created by the keyframe module. 

Through comprehensive experiments we show that \methodname{} is highly effective at producing physically plausible generations, reaching state-of-the-art performance by a significant margin on Physics-IQ Verified~\citep{physicsiq_verified_2026}, a benchmark designed specifically to evaluate physical plausibility. 
Keyframing raises Wan2.2-14B from 32.2 to 55.0 on the Physics-IQ Verified image-to-video track.
The gain also extends to the efficiency of generation: the much lighter Wan2.2-5B reaches 50.0 when supported by our method, substantially outperforming the unmodified 14B backbone at roughly a third of the parameters.
We also report state-of-the-art results on PhyGenBench~\citep{phygenbench_2024}, raising the state of the art from 0.56 to 0.63. 
The two benchmarks test complementary abilities. 
Physics-IQ Verified is image-to-video and scores how a scene evolves from a given observation, so it chiefly measures the motion between frames. 
PhyGenBench is text-to-video and scores whether the frames themselves carry the correct physical appearance of a phenomenon, such as the flame colour of a burning metal, so it chiefly measures accuracy within a frame. 

\methodname{} carries two further advantages over end-to-end pixel generation.
Firstly, a user can inspect and correct the keyframes of a scene before a full video is generated, and so revise an output without incurring the cost of producing a complete sample, which can be very high for modern state-of-the-art video generators.
Secondly, in contrast to pure diffusion, \methodname{} incorporates explicit reasoning into video generation.
The thinking tokens produced by the language model in our keyframe generation module illustrate a consideration of how various objects would interact and move over the course of a video, and lead to correct results in scenes which require reasoning about complex effects such as air pressure or chemical combustion. 

In summary, we make the following contributions:
\begin{itemize}
    \item \methodname{}, a training-free approach to physically plausible video generation which reduces the task posed to a video model to interpolation between keyframes, and which delegates physical planning to a separate module.
    \item A keyframe module in which a vision-language model directs an image edit model, producing a stack of keyframes with timestamps and an expanded caption. Physical reasoning is performed in language, and no physics simulator is used at any stage.
    \item A training-free mechanism for conditioning Wan2.2 on keyframes pinned at arbitrary interior latent frames, which the model's stock conditioning interface cannot express.
    \item State-of-the-art results on Physics-IQ Verified and PhyGenBench, together with an analysis showing that planning effort is spent in proportion to the physical complexity of a scene.
\end{itemize}

\section{Related Work}
\label{sec:related}

\subsection{Boundary- and Keyframe-Conditioned Video Generation}

Conditioning a video model on frames other than the first is an established setting.
Wan2.1~\citep{wan_2025} was later released with a first-last-frame variant, which imposes boundary
conditions at the two ends of the clip, requiring that the model interpolate between these two frames,
and \paper{VACE}{vace_2025}, built on the same backbone, is trained to accept reference frames at other
positions, including randomly chosen ones.
\paper{LTX-Video}{ltxvideo_2024} goes further still, accepting conditioning images at arbitrary frame
positions in its standard inference pipeline.
\citet{wang_2024_inbetweening} adapt an image-to-video model to keyframe interpolation by fine-tuning 
a backwards-in-time variant and combining it with the forward direction.
Meanwhile, \citet{zhu_2024_framewise} drive the same task with explicit frame-wise conditions.
\citet{unitemp_2026} go further, distilling a model that generates in arbitrary
temporal order rather than from a fixed anchor.

These works share our observation that video models can interpolate effectively. 
However, they differ from our work in two respects. First,
they are given their keyframes; these are supplied by an
artist or sampled from a real video, while the contribution is the interpolator. Our
contribution is a system that predicts keyframes. We argue that this is where
physical reasoning has to happen if the video is to be generated. Second, these methods train a dedicated model or adapter for frame conditioning.
We instead condition the stock image-to-video checkpoints without further training, writing keyframes
into interior latent frames through the existing first-frame conditioning channel, at an arbitrary
number of positions chosen in the planning phase according to where the physical motion has specific boundaries.

\subsection{Serial and Parallel Computation}

\citet{serial_scaling_2026} draw a distinction between problems which admit parallel solution and
those which are inherently serial, requiring a chain of dependent steps that cannot be collapsed into
simultaneous work. They show that diffusion models remain within $\mathrm{TC}^0$ however many
denoising steps are taken, so a bidirectional video diffusion model supplies no more serial depth than
its network has layers, irrespective of the length of the clip it generates. Physical evolution is
serial in precisely this sense, since where a scene arrives depends on the states it passed through,
which suggests a reason for the failures measured by the benchmarks above that is independent of data
and parameter count.

One response is to make the generator itself serial, producing a clip in temporal order rather than
all at once. \paper{Diffusion Forcing}{diffusion_forcing_2024} gives each token an independent noise
level, so a causal model can roll a sequence forward beyond its training horizon.
\paper{CausVid}{causvid_2025} distils a bidirectional teacher into a causal student which emits frames
as a stream, and \paper{Self Forcing}{self_forcing_2025} trains such a model on its own rollouts to
close the train-test gap this introduces. \paper{S2PD}{hu_2026} splits the schedule rather than the
architecture, running autoregressive diffusion at high noise, where the serial computation coordinates
interdependent events, before reverting to parallel diffusion at low noise.

These methods buy serial depth with the generator itself: each trains, distils or otherwise modifies
the video model, and generating in temporal order gives up the bidirectional context which
full-sequence diffusion exploits. We obtain the serial computation elsewhere. The keyframe module
advances through a scene one keyframe at a time, each conditioned on the last, so the chain of
physical dependence is resolved serially before any video is generated. What remains for the video
model is a parallel problem, interpolation between states which are already fixed, for which
bidirectional attention over the whole clip is an advantage rather than something to be given up. The
generator is used unmodified and untrained.

\subsection{Agentic and Language-Model-Planned Generation}

A recent line of work combines language models with video generators, and uses the language model to supply the physics which the caption omits. \citet{cect_2026} use a language model to reason
about a chain of causally connected physical events, and subsequently condition a generator on that
chain. \citet{living_worlds_2026} constrain an agent's plan against the capabilities of
a 3D engine and executes it deterministically, thus obtaining physical validity by construction.
\citet{physagent_2026} treat each physical program as an executable hypothesis, closing a loop of simulation, verification and targeted program repair.

Closest to our own system is \paper{NEWTON}{newton_2026}, which likewise uses an agentic planner to
unload physical reasoning from the diffusion process, conditioning the generator on guidance frames and an
expanded caption. The two differ in what they claim. NEWTON is principally concerned with training
its planner, optimised on-policy with Flow-GRPO, and is deliberately generator-agnostic: its keyframes pass
through whatever conditioning interface a frozen generator already exposes. Our claim concerns where the
conditioning enters the generator: Wan2.2's stock interfaces cannot express a keyframe pinned at an interior
latent frame, so we write keyframes into its existing first-frame conditioning channel, and nothing in our
pipeline is trained. Two further differences follow. NEWTON verifies by
scoring generated videos, so each correction cycle costs a full generation, where our agent inspects images
alone; and it writes an independent prompt for each keyframe, where ours are produced by editing the
previous one, which preserves scene consistency.

\subsection{Simulator-Grounded Video Generation}

An alternative is to obtain dynamics from an explicit simulator rather than from a
generative prior. \paper{PhysGen}{physgen_2024} recovers geometry and material properties from a
single image, simulate rigid-body dynamics in image space, and use a diffusion model to
render and refine the result.
\paper{VDAWorld}{vdaworld_2026} takes the abstraction furthest, having a vision-language
model assemble a grounded 2D or 3D scene representation from an image-caption pair,
select a simulator appropriate to the scene, and infer latent dynamics to roll the
state forward. However, it is not a pixel generator: its output is a simulated abstract scene
rather than photorealistic video.

These approaches abstract the scene into a representation a simulator can integrate, leading to
a queryable state and physical guarantees. However, this approach degrades wherever
the abstraction is unable to reproduce accurate dynamics; for example deformation, fluids, fracture and contact-rich interaction. 
We maintain pixels throughout, construct no scene model, and ask
a language model for the states the scene passes through rather than for equations
defining them. A recent survey situates both lines of work within a common taxonomy of physical cognition in video generation~\citep{physics_cognition_2025}.

\begin{figure*}[!t]
    \centering
    \includegraphics[width=0.85\linewidth]{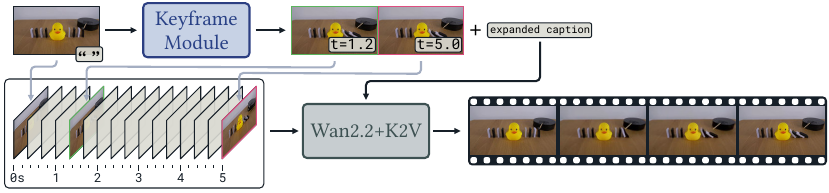}
    \caption{%
        Overview of the proposed method. We design a keyframe module which reasons about the physical evolution of a given scene and produces keyframes with timestamps and an expanded caption. These are then used as the conditioning signal for Wan2.2, resulting in videos with more accurate physical plausibility.
    }
    \label{fig:overview}
\end{figure*}

\subsection{Measuring Physical Plausibility}

Physics-IQ Verified~\citep{physicsiq_verified_2026} is our primary benchmark, and reports the finding that
motivates this work: visual quality and physical understanding are largely uncorrelated
across current generators. 
VideoPhy-2~\citep{videophy2_2025} evaluates action-centric
physical commonsense, and PhyGenBench~\citep{phygenbench_2024} scores 160 prompts across
27 physical laws in four domains. 
Both these are used by the agentic methods discussed above,
and we report on PhyGenBench for direct comparison. 
\citet{mechanistic_2026} separate
what a video model represents about scene state from what it propagates as dynamics, a
distinction that our division of labour enforces in its architecture: the planner is responsible
for state at chosen instants, while the video model is a stateless interpolant of the motion which happens between those keyframes.

Several image-to-video entries were added to the Physics-IQ Verified leaderboard~\citep{physicsiq_leaderboard_2026} very recently, the strongest being FLUX~3 [large] (51.1), Odyssey~3 Pro (50.0), Odyssey~3 with prompt enhancement (48.8), and Physis-Lang applied to Cosmos3-Super and Cosmos3-Nano (48.2 and 43.3).
We treat these as concurrent work and do not include them in Table~\ref{tab:leaderboard}, but \methodname{} with Wan2.2-14B (55.0) nonetheless outperforms all of them with a single generation per scene. It also scores above the best-of-eight submissions of FLUX~3 [large] and Odyssey~3 Pro (54.7 each), which generate eight candidates per scene and select one with a reward model or a ranking.

\section{Method}

Our approach leverages a generative keyframing module to produce keyframes along with timestamps, marking out distinct physical phases which occur in the generated video.
This keyframe set is used as conditioning for a video generation model. 
In addition, the keyframe generation module returns an expanded caption, which describes precisely how the scene evolves, further reducing the physical inference which the diffusion model must perform itself. 
An overview of our method is shown in Figure~\ref{fig:overview}.

\subsection{Keyframe Module}
\label{sec:keyframe_module}

\begin{figure*}[tb]
    \centering
    \includegraphics[width=0.85\linewidth]{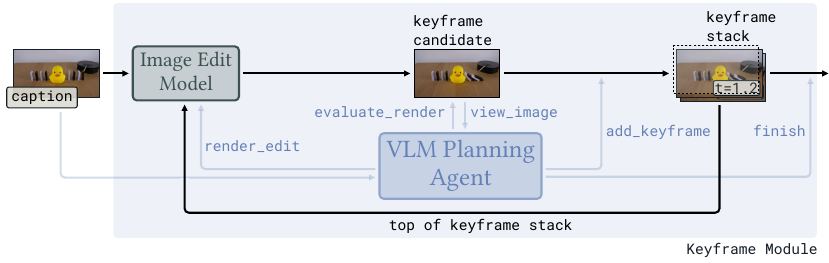}
    \caption{%
        Keyframe module detail. A VLM planning agent orchestrates a recursive loop producing keyframes. The first frame is passed to an image editor with an instruction from the VLM to produce the appearance of the frame later in time. The VLM reviews this frame and, if deemed correct, the frame is added to a stack of keyframes. The VLM repeats this process, now passing the top image to the image editor to add to the keyframe stack. Once all keyframes are produced, the model finishes, returning an expanded caption describing the motion of objects in the video with the keyframe stack. 
    }
    \label{fig:keyframe_detail}
\end{figure*}

The keyframe module is an agentic pipeline. 
A central planning agent, \paper{Gemini 3.8 Flash}{deepmind_2026}, orchestrates the keyframe generation. 
The planning agent is given the original caption for the video, a task-agnostic prompt, and, optionally, the image intended for the video generation model. 
Where no image is supplied, as in the text-conditioned setting, the agent generates the first keyframe from the caption alone, and generation proceeds identically thereafter. 
The prompt informs the agent that it has a set of tools which it should use to generate a stack of keyframes which represent the motion in the next five seconds of video. 
The general pipeline by which the planning agent generates the keyframes is fixed, although it is given flexibility through its use of tooling. 
A diagram of the keyframe module is shown in Figure~\ref{fig:keyframe_detail}. 

First, the planning agent is directed to imagine how the scene will evolve in the next five seconds, and consider how these could be divided by keyframes into distinct phases of motion. 
The agent generates each keyframe in sequence, enforcing the causal nature of physically-plausible generation. 

For each keyframe, the agent writes an editing instruction describing how the scene should change to reach the next keyframe.
This instruction is passed to an image edit model, \paper{Gemini 3.1 Flash Image}{deepmind_2026b}, along with the most recently-generated keyframe.
If no keyframes have been generated yet, the input image is passed to the image edit model. 
Once the keyframe is generated, it is added to the stack of keyframes along with a timestamp for the keyframe provided by the planning agent. 

\subsubsection{Agentic Tooling}

Tooling gives flexibility to this generation loop. 
At each stage, the agent is free to select which tool it uses, so the pace of the progression through the keyframe generation pipeline is controlled by the agent, and many loops may be undertaken until the keyframes are asserted to be complete.
The tools let the agent render an edit, view and evaluate the result, commit it to the keyframe stack with a timestamp, and close the loop with an expanded caption. 
Their full specifications are given in Appendix~\ref{sec:tools}.

\subsubsection{Repeat Keyframes}
\label{sec:repeat}

It is often advantageous for the model to repeat a keyframe where no motion occurs in a given time interval, which it can do trivially by calling \texttt{add\_keyframe(time)} twice with no \texttt{render\_edit(instruction)} between them. 
However, often, when two identical keyframes are placed far apart in time, for example at $0.0s$ and at $5.0s$, the video model invents spurious motion between these timestamps. 
For this reason, where repeat keyframes are used, we also automatically insert additional keyframes every second between the two identical keyframes. 
We additionally hold the video still once the plan says the scene has come to rest. 
Where the plan ends in repeated keyframes, or where its final two keyframes are near-identical (a mean absolute difference below 4 grey levels), we repeat the frame at the start of that interval for the remainder of the video. 
We ablate this treatment of repeated keyframes in Section~\ref{sec:ablations}.

\subsection{Conditioning a Video Generation Model With Keyframes}

For our video generation model, we use Wan2.2 I2V-A14B and \paper{Wan2.2 TI2V-5B}{wan_2025}. 
By default, Wan2.2 accepts only an initial conditioning frame and caption pair. 
However, we observe that it is trivially possible to condition Wan2.2 for multiple frames at several points in a given video. 

\subsubsection{First Frame Conditioning Wan2.2}

Conventionally, Wan2.2 is conditioned with a single image $I\in \mathbb{R}^{H\times W \times 3}$. 
To perform the conditioning, this image is concatenated with $N$ additional frames, which are all set to be mid-grey (N.B. mid-grey eventually maps to $\mathbf{0}$ since the tensor is normalised to $[-1,1]$ in the VAE), producing a padded image tensor of $N+1$ frames:

\begin{equation}
    I'[n] =
\begin{cases}
I & n = 0\\
\mathbf{0.5} & \text{otherwise}
\end{cases}\in \mathbb{R}^{(N+1)\times H\times W \times 3}
\end{equation}

The padded image tensor $I'$ is passed through the VAE of Wan2.2. 
This VAE changes the dimension of the tensor to $(N/L_T+1)\times H/L_H \times W/L_W \times C$ where $L_T, L_H, L_W$ are all hyperparameters of the VAE relating to spatial and temporal compression, and $C$ is the feature dimension of the VAE.

A mask vector is constructed which is simply $1$ repeated $L_T$ times followed by $N$ zeros to indicate that only the first frame is filled. 
The mask vector is reshaped to $(N/L_T+1)\times L_T$. 
This is then concatenated in the channel dimension with the VAE output, resulting in a final conditioning tensor with shape $(N/L_T+1)\times H/L_H \times W/L_W \times (F+L_T)$.

The full conditioning tensor is concatenated with the noise tensor along the channel dimension before being passed to the denoising transformer.
At each subsequent denoising step, the conditioning tensor is re-concatenated to the noise tensor before passing to the transformer.

\subsubsection{Keyframe Conditioning Wan2.2}

Our approach is a trivial extension of this. 
We simply replace the grey frame at each keyframe timestamp with the corresponding generated keyframe. 
Given a set of $M$ keyframes $K_i\in \mathbb{R}^{H\times W\times 3}, 0\leq i <M$ with associated timestamps $t_i$ we construct the new padded image-keyframe mixture:

\begin{equation}
    I'[n] =
\begin{cases}
I & n = 0\\
K_i & n = a_i,\ 0 \le i < M\\
\mathbf{0.5} & \text{otherwise}
\end{cases}
\end{equation}

Where the indices of the placed keyframes are set to be the nearest frame which corresponds to the provided timestamp at the end of one latent timestep (i.e. every $L_T$ frames), and clamped to be between frame number $L_T$ and the final frame. 

\begin{equation}
a_i = L_T\left\lfloor \frac{f}{L_T} t_i \right\rceil ,\qquad a_i \in \{L_T, 2L_T, \dots, N\}
\label{eqn:keyframe_arg}
\end{equation}

Where $f$ is the frame rate of the generator in frames per second.

The reason that frames are only inserted every $L_T$ frames is because we note in our experiments that intermediate conditioning only works if the conditioning is applied at the end of a temporal block of $L_T$ frames. 
A single temporal block represents the set of frames Wan2.2's VAE treats as a single latent frame, and our hypothesis is that by placing a conditioning frame at the end of such a block, Wan2.2 is able to treat it most similarly to an initial conditioning frame. 

Finally, we update the mask vector to indicate that the specific keyframe slots are now also filled.

\begin{figure*}[tb]
    \centering
    \includegraphics[width=0.85\linewidth]{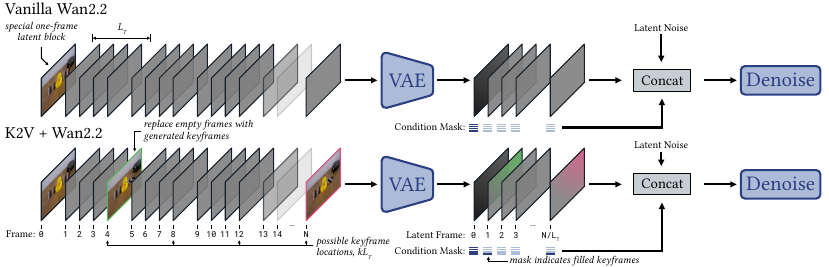}
    \caption{%
        Conditioning model detail. Keyframes are inserted into the typical conditioning stack of blank frames at locations designated in Equation~\ref{eqn:keyframe_arg}. This is then passed to the VAE, and is subsequently treated as normal conditioning in the Wan pipeline. The conditioning is concatenated with a mask indicating which keyframes are filled, and with the latent noise before being passed to the transformer. The mask and latent conditioning is appended before each denoising step. 
    }
    \label{fig:conditioning_detail}
\end{figure*}

\FO{Surprisingly, }this approach works with no training or fine tuning of Wan2.2, and works with the default weights. 
This appears to be a surviving behaviour of the Wan2.2 network from its pre-training, which did explicitly include frame interpolation as one task~\cite{wan_2025}.
In all our experiments we use $N=80$, while the Wan2.2 VAE has temporal compression factor $L_T=4$ and frame rate $f=16$. This approach is illustrated in Figure~\ref{fig:conditioning_detail}

\subsection{Image and Video Post-Processing}
\label{sec:postprocessing}

While the method works remarkably well out of the box, we notice improvements to our results with the use of image and video post-processing.
We attribute this to the fact that the image edit model, while very successful at maintaining physical consistency, occasionally introduces subtle color variations relative to the edit image.
For this reason, before passing the keyframes to the video model, the images are colour matched using a simple affine transformation to the input image. 
Once the video is generated, we also match each frame to the first frame in the same manner. 
This mitigates any artefacts induced by the subtle colour variations resulting from the image edit model.

\section{Results}

We select our experiments to show the broad applicability of our approach in first frame and text conditioned settings. 
Our results highlight that, both qualitatively and quantitatively, our approach outperforms existing methods and produces videos which correctly follow physical laws. 

\subsection{First Frame-Conditioned Generation}

To examine our approach on first frame-conditioned video generation, we use the Physics-IQ Verified benchmark. 
This benchmark is designed to evaluate video models' ability to adhere to physical laws across a range of categories. 
Categories include solid mechanics, optics, thermodynamics, magnetism and fluid mechanics. 
In each example, two ground truth videos are provided within Physics-IQ Verified from a single identical setting, and generative models are scored on whether the variation between their generation and one of the ground truth videos is greater than the variance between the two ground truth samples. 
Thus, all results are scored as an error beyond physical variance. 
We evaluate our method on the image to video setting, wherein the first frame of a video and a caption describing the setting is passed to the video generator. 

\begin{figure*}[tb]
    \centering
    \includegraphics[width=\linewidth]{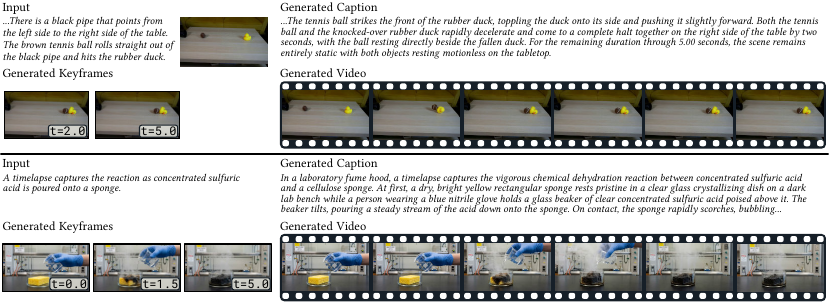}
    \caption{%
        Qualitative results from our approach. The top row shows an example from Physics-IQ Verified and the bottom row an example from PhyGenBench. These results highlight the diverse placement of keyframes in different videos, and the physical quality of the resulting generated videos. 
    }
    \label{fig:qualitative_1}
\end{figure*}

We show a selection of results in Figure~\ref{fig:qualitative_1}. 
Our approach produces videos which strictly adhere to physical laws such as gravity, collisions, and which preserve the number and size of objects throughout the video. 

Quantitatively, our approach outperforms all baselines by a significant margin. 
Compared to a vanilla implementation of Wan2.2-14B, our approach leads to an improvement of 22.8 points. 
Compared to the prior state of the art, our approach leads to an improvement of 12.3 points.
We attribute the significant improvement in this metric to the ability of our method to reason about and plan the physical evolution of a scene, and to its ability to impose that plan on the video generation phase. 

\begin{table}[t]
\centering
\caption{%
  Physics-IQ Verified, image-to-video track. Higher is better and 100 is the maximum possible score. Methods marked with $^\dag$ are ours.
}
\label{tab:leaderboard}
\begin{tabular}{rlc}
\toprule
& Model & Physics-IQ Verified \\
\midrule
1 & \methodname+Wan2.2-14B$^\dag$               & \cellcolor{bestcell} 55.0 {\small $\pm$ 0.8} \\
2 & \methodname+Wan2.2-5B$^\dag$           & \cellcolor{secondcell} 50.0 {\small $\pm$ 1.3} \\
3 & Cosmos3-Super                            & 42.7 {\small $\pm$ 0.8} \\
4 & MiniMax H3 {\small (FL2VA)}              & 39.8 {\small $\pm$ 0.4} \\
5 & Cosmos3-Nano                             & 37.3 {\small $\pm$ 0.9} \\
6 & MiniMax H3 Max                           & 36.2 {\small $\pm$ 0.7} \\
7 & Grok Imagine Video                       & 34.8 {\small $\pm$ 0.6} \\
8 & Magi-1 24B + GeoPhys (BoN) {\small (op)} & 33.7 {\small $\pm$ 1.4} \\
9 & Hunyuan Video 1.5                        & 33.4 {\small $\pm$ 0.8} \\
10 & Cosmos3-Edge                            & 32.7 {\small $\pm$ 1.1} \\
11 & Wan2.2-14B & 32.2 {\small $\pm$ 0.6} \\
12 & CogVideoX-5B                            & 31.8 {\small $\pm$ 1.5} \\
13 & Kandinsky-WM 1.0                        & 30.8 {\small $\pm$ 0.9} \\
14 & Magi-1 24B {\small (op)}                & 30.2 {\small $\pm$ 1.1} \\
15 & Wan2.2-TI2V-5B & 27.7 {\small $\pm$ 0.9} \\
16 & Sora 2                                  & 26.5 {\small $\pm$ 0.8} \\
17 & P-Video                                 & 25.3 {\small $\pm$ 1.8} \\
\bottomrule
\end{tabular}

\end{table}

\begin{figure*}[tb]
    \centering
    \includegraphics[width=\linewidth]{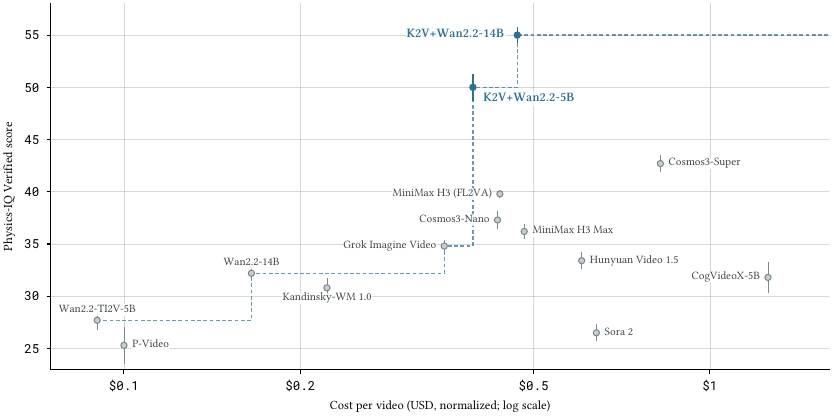}
    \caption{%
        Physics-IQ Verified score against cost per video for the image-to-video models of Table~\ref{tab:leaderboard}, with costs as listed on the Physics-IQ Verified leaderboard~\citep{physicsiq_leaderboard_2026} on 8 October 2026.
        Costs follow the leaderboard's normalization, which scales generation cost to 24\,fps and 1280-pixel-wide output and then adds any language-model cost; for \methodname{} this is the leaderboard's Wan2.2 generation cost plus our planning cost from Table~\ref{tab:cost}.
        The dashed line traces the cost score frontier, and vertical bars show the standard deviations of Table~\ref{tab:leaderboard}.
    }
    \label{fig:cost_frontier}
\end{figure*}

We also show that our approach achieves this result while maintaining a moderate cost. 
Figure~\ref{fig:cost_frontier} places \methodname{} alongside the models of Table~\ref{tab:leaderboard} on the cost score chart published with the leaderboard~\citep{physicsiq_leaderboard_2026}.
Under the leaderboard's own normalization, \methodname{} with Wan2.2-14B costs \$0.47 per video and scores 55.0, and with Wan2.2-5B it costs \$0.39 and scores 50.0.
Both lie on the cost score frontier, and each scores higher than every other model in the table at any cost.
A full breakdown of where this cost is incurred is given in Appendix~\ref{sec:cost}.

\begin{figure*}[tb]
    \centering
    \includegraphics[width=\linewidth]{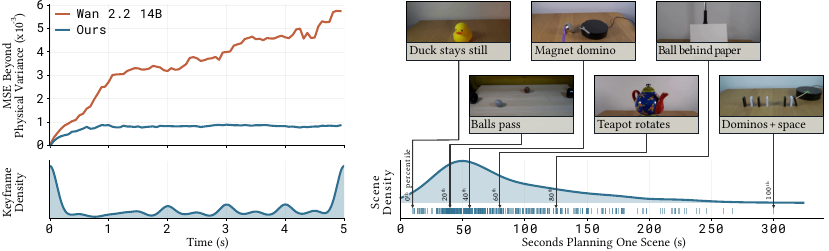}
    \caption{%
        \textbf{Left: } Compared to vanilla Wan2.2, the quality of our generations (measured by mean squared error above physical variance) does not rise over the 5 seconds of a video. Beneath, keyframe density shows that our approach distributes keyframes relatively evenly throughout videos on Physics-IQ Verified, with spikes at 0s and 5s, as these keyframes are required for each video. \textbf{Right: } The time taken to produce keyframes shows that the language model spends compute efficiently, spending more time on scenes which require more careful planning and less time on simpler scenes.
    }
    \label{fig:analysis_1}
\end{figure*}

As well as scoring more highly, we also highlight the advantage of our method that, as opposed to pure video diffusion, the error made by our method does not appear to drift over the course of the video. 
We show this in the upper-left panel of Figure~\ref{fig:analysis_1}. 
Whereas generations from Wan2.2-14B degrade over the course of a video, straying further and further from the ground truth, our approach does not. 
Instead, its mean squared error beyond physical variance remains fairly consistent over the examined 5 seconds.

We additionally highlight that, because our method effectively implements reasoning via its use of a language model, the approach efficiently spends compute on those scenes and moments which are particularly challenging. 
All videos are required by the prompt to use keyframes at 0s (the input image) and at 5s; but beyond these constraints the model is free to add keyframes where it thinks they are required. 
The keyframe density plot shown in the lower-left panel of Figure~\ref{fig:analysis_1} highlights that the model does use keyframes at a range of moments within videos. 

The right panel of Figure~\ref{fig:analysis_1} shows that the language model also spends more time on scenes which are particularly complex. 
A trivial scene of a duck remaining motionless is reasoned about for less than 10 seconds, whereas a complex scene showing a row of dominos with a gap in the middle takes more than 5 minutes, since the model must reason about whether the domino chain will cross the gap. 
The reasoning trace extracted from the language model for this clip reveals the model considering this problem: \emph{``I can see that the gap between domino 4 and domino 5 is wide enough for the domino 5 to fall without hitting domino 4 and for it to be a valid demonstration''.}

\subsection{Text-Conditioned Generation}

We evaluate text-conditioned generation on the \paper{PhyGenBench}{phygenbench_2024} benchmark. 
In PhyGenBench, text-to-video models are required to generate videos of different physical phenomena across settings which span a wide range of domains, such as mechanics, optics, thermodynamics and materials. 

Models are evaluated on their ability to accurately visualise the features which uniquely determine the physical phenomena, such as showing the correct flame color of copper burning, or correctly reasoning about the buoyancy of objects of different densities. 

\begin{figure*}[p]
    \centering
    \includegraphics[width=0.68\linewidth,trim={0 245.64pt 0 0},clip]{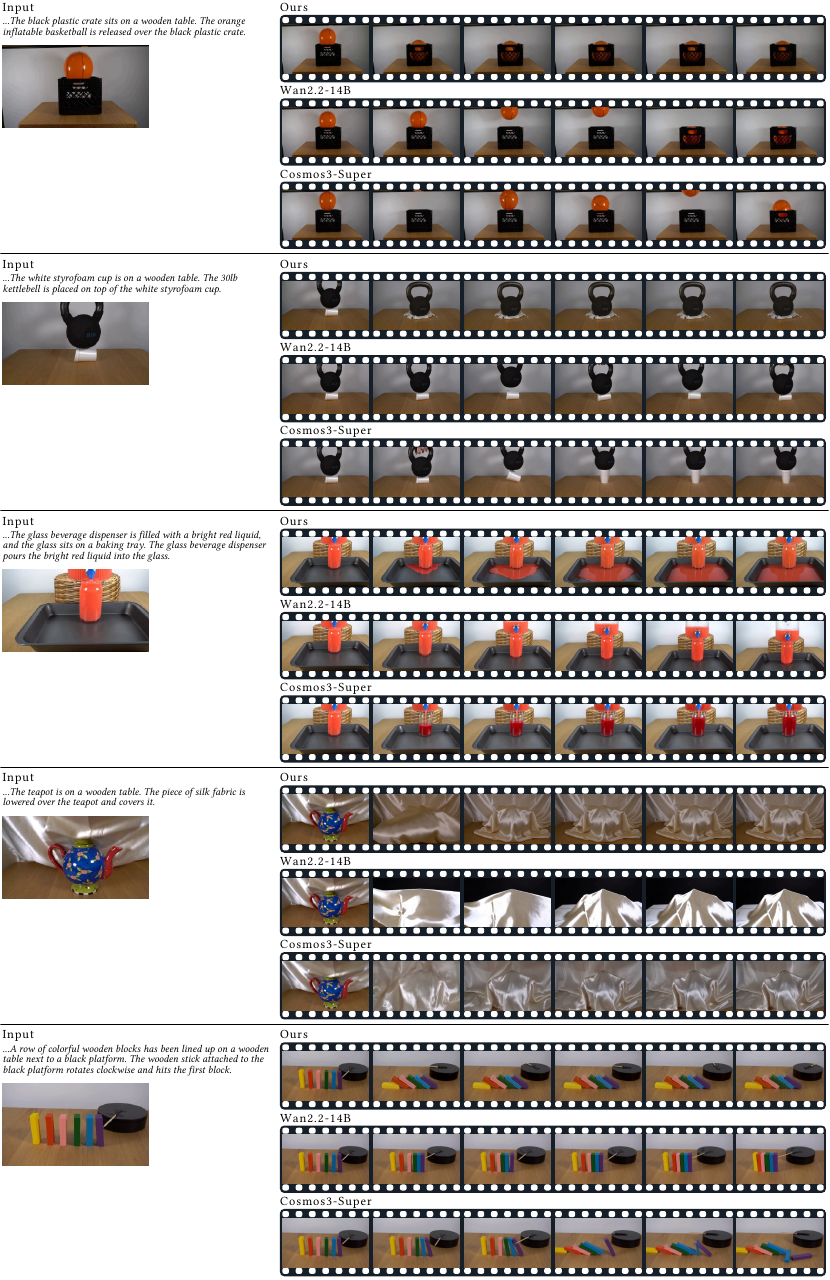}\\[4pt]
    \includegraphics[width=0.68\linewidth]{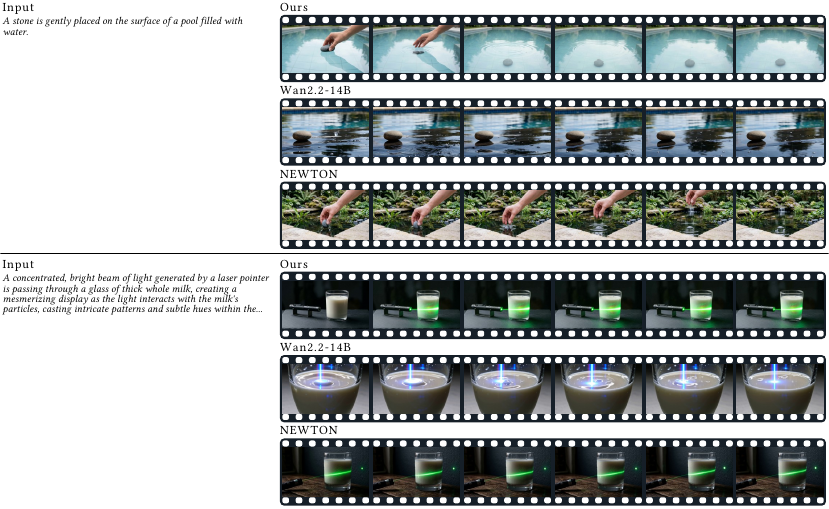}
    \caption{%
        Qualitative comparison. \textbf{Top: } Physics-IQ Verified (centre view), where ours, Wan2.2-14B and Cosmos3-Super are conditioned on the same input frame. \textbf{Bottom: } PhyGenBench, where ours, Wan2.2-14B and NEWTON\protect\footref{fn:opennewton} are given the same text prompt and no image. Frames are shown at 1\,s intervals. Cosmos3-Super is image-to-video only and so does not appear for PhyGenBench. Further Physics-IQ comparisons are given in Figure~\ref{fig:appendix_comparison}.
    }
    \label{fig:qualitative_comparison}
\end{figure*}

Qualitatively, we show in Figures~\ref{fig:qualitative_1} and~\ref{fig:qualitative_comparison} that the language reasoning enforces correct physics throughout our generation.

Figure~\ref{fig:qualitative_comparison} compares two PhyGenBench prompts against Wan2.2-14B and NEWTON\footnote{The full code for NEWTON~\citep{newton_2026} has not been released. Its authors provide OpenNEWTON, a version which uses no trained components, and which we run with its default GPT-5.5 planner and Seedance 2.0 Fast video generator. All NEWTON videos and costs reported in this paper are produced with it.\label{fn:opennewton}}.
For a stone placed on the surface of a pool, Wan2.2-14B rests the stone on a surface it invents inside the pool, and NEWTON replaces the pool with a garden pond in which the stone never visibly sinks, while in ours the stone sinks to the floor of the pool.
Thick whole milk should scatter a laser beam along its whole path, as it does in ours, where the entire glass glows; NEWTON's beam instead crosses the glass as a clean line and emerges as a sharp spot on the wall behind it, which is impossible in a liquid as cloudy as milk, and Wan2.2-14B shows a beam striking the surface from above with little scattering.

This is supported quantitatively. 
Table~\ref{tab:phygenbench} shows that our approach outperforms all other methods, recording state of the art performance. 

\begin{table*}[t]
\centering
\caption{%
  PhyGenBench, text-to-video. Scores are PCA from PhyGenEval, averaged over the two
  judge tracks as the benchmark specifies; higher is better and 1.0 is the ceiling.
  Methods marked $^\dag$ are ours.
  }
\label{tab:phygenbench}
\begin{tabular}{rlccccc}
\toprule
& Method & Mech. & Optics & Therm. & Mat. & Avg. \\
\midrule
1  & \methodname+Wan2.2-14B$^\dag$        & \cellcolor{secondcell} 0.583 & \cellcolor{secondcell} 0.707 & \cellcolor{bestcell} 0.589 & \cellcolor{bestcell} 0.625 & \cellcolor{bestcell} 0.633 \\
2  & \methodname+Wan2.2-5B$^\dag$         & \cellcolor{bestcell} 0.600 & \cellcolor{bestcell} 0.713 & \cellcolor{secondcell} 0.578 & \cellcolor{secondcell} 0.608 & \cellcolor{bestcell} 0.633 \\
3  & LTX-Video-2B~\citep{ltxvideo_2024} + NEWTON & 0.500 & 0.647 & 0.522 & 0.542 & 0.560 \\
4  & Wan2.2-TI2V-5B                 & 0.550 & 0.580 & 0.533 & 0.500 & 0.544 \\
5  & Wan2.2-T2V-A14B               & 0.500 & \cellcolor{secondcell} 0.707 & 0.433 & 0.400 & 0.527 \\
6  & LTX-Video-2B~\citep{ltxvideo_2024} & 0.508 & 0.580 & 0.478 & 0.450 & 0.510 \\
7  & Gen-3                                  & 0.45 & 0.57 & 0.49 & 0.51 & 0.51 \\
8  & Kling                                  & 0.45 & 0.58 & 0.50 & 0.40 & 0.49 \\
9 & Vchitect 2.0 {\small (2B)}           & 0.41 & 0.56 & 0.44 & 0.37 & 0.45 \\
10 & CogVideoX-5B                           & 0.43 & 0.55 & 0.40 & 0.42 & 0.45 \\
11 & Open-Sora V1.2 {\small (1.1B)}       & 0.43 & 0.50 & 0.44 & 0.37 & 0.44 \\
12 & Pika                                   & 0.35 & 0.56 & 0.43 & 0.39 & 0.44 \\
13 & Cosmos-Predict2.5                & 0.300 & 0.333 & 0.411 & 0.385 & 0.352 \\
\bottomrule
\end{tabular}
\end{table*}

\methodname{} costs approximately \$0.47 per video on Physics-IQ Verified and \$0.59 on PhyGenBench, comprising \$0.30 and \$0.42 of planning respectively and \$0.165 for the Wan2.2-14B generation (Appendix~\ref{sec:cost}).
The publicly-available version of NEWTON, OpenNEWTON\footref{fn:opennewton} costs approximately \$3 per video on PhyGenBench prompts, because its verifier scores complete generations: it rendered four 5-second clips per prompt on average, where \methodname{} renders one and verifies on still images.

\subsection{Controllability}

The right panel of Figure~\ref{fig:analysis_2} shows that the keyframes are also trivially controllable via user-specified edits. 
In this example, two keyframes are generated of a flower bud, and the flower once blossomed using \paper{Gemini 3.1 Flash Image}{deepmind_2026b}. 
These are used to produce a video showing the full timelapse of the bud opening.
We then use the same image model to simply update the final image, changing the color of the inner petals to be white and orange. 
Regenerating the video with these keyframes correctly updates the video dynamics.

\subsection{Keyframe Conditioning}

\begin{figure*}[tb]
    \centering
    \includegraphics[width=\linewidth]{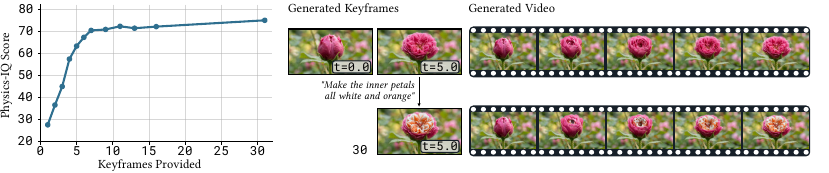}
    \caption{%
        \textbf{Left: } Conditioning Wan2.2-5B on $N$ equally spaced ground truth keyframes, for $N$ between 1 and 6, raises the Physics-IQ Verified score from 27.7 to 67.4. \textbf{Right: } Editing the final keyframe with the same image edit model used in our pipeline yields a new video, here with the inner petals recoloured.
    }
    \label{fig:analysis_2}
\end{figure*}

We next verify that conditioning on keyframes is what drives the improvement, independently of how well the keyframes are chosen.
In this experiment, we take $N$ evenly-spaced keyframes from ground-truth videos in the Physics-IQ Verified dataset. 
We then condition Wan2.2-5B with one trial on these ground-truth keyframes in the central perspective and score the resulting videos. 
Our results show that this conditioning structure consistently improves Physics-IQ Verified score as the number of keyframes increases.
Results are shown in the left panel of Figure~\ref{fig:analysis_2}.

While increasing the number of keyframes initially linearly improves the Physics-IQ Verified score, it plateaus where a high number of keyframes are used, indicating that once the motion is specified by the supplied keyframes, adding information in the form of more keyframes does not improve generation. 
We also note that in this experiment the keyframes are selected to be evenly-spaced, and are therefore naive in their placement. 
We propose to use a small number of keyframes at carefully placed locations within a given video. 

\subsection{Ablations}
\label{sec:ablations}

We conduct a thorough ablation of all components of our method on the central perspective of Physics-IQ Verified. 
We ablate the agentic loop, instead prompting the language model once to provide a chain of image edit instructions which are passed directly to the image model. 
This reveals that our agentic pipeline is essential due to its ability to review and correct errors made by the image edit model, or to refine its image edits. 
We also replace the language and image editing models with open-weight variants, using \paper{Qwen3-VL Instruct}{qwen3_vl} for planning and the recently released \paper{Qwen-Image-2.1}{qwen_image21} for image editing.
This variant scores 40.2, 15.8 points below the full method, which shows that closed-source models are presently still needed for their ability to edit an image with the precision required to construct an accurate video.

We additionally ablate several conditioning features of our method. 
We ablate any keyframes provided which do not sit at 0s or 5s, and show that intermediate keyframes are essential for producing accurate generations. 
We ablate the conditioning to be sparse between identical keyframes, so that when two keyframes are provided which are identical, we do not automatically repeat this identical keyframe every second between the two keyframes, nor hold the video still once the plan declares the scene at rest, both as described in Section~\ref{sec:repeat}.
This demonstrates our earlier note, that the video model often invents spurious motion betwen two identical keyframes. 
We ablate the benchmark caption, instead providing the original caption which was provided with the input frame, which shows that densely prompting the motion in a scene also supports more accurate video generation. 

Finally, we ablate color correction, showing that the use of keyframes does lead to color varying throughout a video, but that this can be rectified with a simple affine transformation of the frame colors.

\begin{table}[t]
\centering
\caption{%
  Ablations on Physics-IQ Verified, centre perspective. $\Delta$ is measured against the full method at the top.
}
\label{tab:ablations}
\begin{tabular}{lcc}
\toprule
& Physics-IQ & $\Delta$ \\
\midrule
\textbf{\methodname{} (full)}              & \textbf{56.0} & --- \\
\midrule
\multicolumn{3}{l}{\emph{Planning}} \\
\quad no agentic loop                      & 54.3 & $-1.7$ \\
\quad open-weight models (Qwen)       & 40.2 & $-15.8$ \\
\midrule
\multicolumn{3}{l}{\emph{Conditioning}} \\
\quad endpoints only                       & 48.6 & $-7.5$ \\
\quad sparse conditioning                & 51.9 & $-4.1$ \\
\quad original caption                    & 53.9 & $-2.1$ \\
\midrule
\emph{No Colour correction}                   & 52.7 & $-3.3$ \\
\bottomrule
\end{tabular}
\end{table}

\section{Limitations}

Our proposed approach incurs a minor additional cost to the process of generation. 
On average for Physics-IQ Verified, the median time taken to plan one scene was 62 seconds. 
However, we highlight here that this is a minor cost compared to the time required to generate a video, which is around 32 minutes for Wan2.2-14B. 
We additionally note that scaffolding Wan2.2-5B, which requires just 4 minutes per generation, with our method outperforms the 14B model, leading to an effective acceleration of accurate generation. 

Although our method does resolve almost all physical plausibility problems in our test sets, there remain some instances where the language model fails to reason correctly about the physics of a scene, leading to an incorrect generation. 
For example, in one scene a stack of blocks was reasoned incorrectly to be stable, whereas in fact the blocks were visibly unbalanced. This led to the generation of incorrect keyframes, and ultimately an implausible video. 
We expect that the use of a stronger language model would further improve results. 

Finally, while our method can set moments in a video, it cannot currently determine intermediate motion. 
This can lead to remaining physical implausibilities where the model inserts incorrect motion. 
We observe this most strongly in scenes with balls rolling, where a ball follows a trajectory which is not straight between keyframes, violating Newton's First Law of motion. 

\section{Conclusions and Discussion}

In conclusion, we proposed \methodname{}, a novel approach for video generation which enforces adherence to set keyframes which are generated with a custom language model-backed keyframe module. 
We proposed our method to redistribute the physical generation task required for accurate video generation from within video diffusion onto a language model, which has the ability to reason in text and to plan how a scene will evolve physically. 

Through our experiments we show that our approach is significantly more successful at generating videos which are physically plausible than any existing video generators. 
On the Physics-IQ Verified benchmark we demonstrate state-of-the-art results by a significant margin. 
Our approach scaffolding Wan2.2-14B outperforms Wan in its vanilla state by 22.8 points; and outperforms the existing state of the art by 12.3 points. 
On PhyGenBench our method outperforms all comparable approaches in all physical categories tested.

\section{AI Disclosure}

In this work, we used generative AI tools for implementing our approach in code and for providing feedback on research methodology. 
We additionally used generative AI tools for suggesting experimental parameters and proposing related work.
We have reviewed all AI-assisted work, and take responsibility for the final content of this work.

\section*{Ethics and Privacy Statement}

The rapid proliferation of generative video models has had widespread effects. 
These include significant beneficial progress in fields such as computer graphics, robotics, and physical intelligence. 
However, some effects have been negative, such as the distribution of content intended to deceive viewers. 
Research which intends to improve the physical plausibility of generated content could make detection of generated content more difficult; 
and developments in this field should be balanced with work to facilitate detection and watermarking of generated content. 
However, presently we feel that the benefits of this research mean that improving physical plausibility is a justified objective.

\begin{acks}
    We thank Jeffrey Hu for reviewing the manuscript and providing valuable feedback. 
    Research supported by the NVIDIA Academic Grant Program using NVIDIA RTX PRO 6000 Blackwell GPUs. This work was funded in part by Toyota Motor Europe.
\end{acks}

\bibliographystyle{ACM-Reference-Format}
\bibliography{bibliography}

@inproceedings{serial_scaling_2026,
      title={The Serial Scaling Hypothesis}, 
      author={Yuxi Liu and Konpat Preechakul and Kananart Kuwaranancharoen and Yutong Bai},
      year={2026},
      booktitle={International Conference on Learning Representations (ICLR)},
      eprint={2507.12549},
      archivePrefix={arXiv},
}

@inproceedings{diffusion_forcing_2024,
      title={Diffusion Forcing: Next-token Prediction Meets Full-Sequence Diffusion}, 
      author={Boyuan Chen and Diego Marti Monso and Yilun Du and Max Simchowitz and Russ Tedrake and Vincent Sitzmann},
      year={2024},
      eprint={2407.01392},
      archivePrefix={arXiv},
      booktitle={Advances in Neural Information Processing Systems (NeurIPS)},
}

@inproceedings{causvid_2025,
  title={From Slow Bidirectional to Fast Autoregressive Video Diffusion Models},
    author={Yin, Tianwei and Zhang, Qiang and Zhang, Richard and Freeman, William T and Durand, Fredo and Shechtman, Eli and Huang, Xun},
  booktitle={Proceedings of the IEEE/CVF Conference on Computer Vision and Pattern Recognition (CVPR)},
  year={2025},
  eprint={2412.07772},
  archivePrefix={arXiv},
}

@inproceedings{self_forcing_2025,
  title={Self Forcing: Bridging the Train-Test Gap in Autoregressive Video Diffusion},
      author={Xun Huang and Zhengqi Li and Guande He and Mingyuan Zhou and Eli Shechtman},
  booktitle={Advances in Neural Information Processing Systems (NeurIPS)},
  year={2025},
  eprint={2506.08009},
  archivePrefix={arXiv},
}

@misc{hu_2026,
      title={S2PD: Serial-to-Parallel Diffusion for Physically and Logically Consistent Video Generation}, 
      author={Jeffrey Hu and Daniel Olmeda Reino and Ayush Tewari},
      year={2026},
      eprint={2610.06847},
      archivePrefix={arXiv},
      primaryClass={cs.CV},
      url={https://arxiv.org/abs/2610.06847}, 
}

@InProceedings{motamed_2026,
    author={Motamed, Saman and Culp, Laura and Swersky, Kevin and Jaini, Priyank and Geirhos, Robert},
    title={Do Generative Video Models Understand Physical Principles?},
    booktitle={Proceedings of the IEEE/CVF Winter Conference on Applications of Computer Vision (WACV)},
    year={2026}
}

@misc{deepmind_2026, title={Gemini 3.8 Flash - model card}, url={https://deepmind.google/models/model-cards/gemini-3-8-flash/}, journal={Google DeepMind}, author={Google DeepMind}, year={2026}, month={Sep}}

@misc{deepmind_2026b, title={Gemini 3.1 Flash Image (Nano Banana 2)}, url={https://docs.cloud.google.com/gemini-enterprise-agent-platform/models/gemini/3-1-flash-image}, journal={Google Cloud}, author={Google DeepMind}, year={2026}, month={Oct}}

@misc{wan_2025,
      title={Wan: Open and Advanced Large-Scale Video Generative Models}, 
      author={Team Wan and Ang Wang and Baole Ai and Bin Wen and Chaojie Mao and Chen-Wei Xie and Di Chen and Feiwu Yu and Haiming Zhao and Jianxiao Yang and Jianyuan Zeng and Jiayu Wang and Jingfeng Zhang and Jingren Zhou and Jinkai Wang and Jixuan Chen and Kai Zhu and Kang Zhao and Keyu Yan and Lianghua Huang and Mengyang Feng and Ningyi Zhang and Pandeng Li and Pingyu Wu and Ruihang Chu and Ruili Feng and Shiwei Zhang and Siyang Sun and Tao Fang and Tianxing Wang and Tianyi Gui and Tingyu Weng and Tong Shen and Wei Lin and Wei Wang and Wei Wang and Wenmeng Zhou and Wente Wang and Wenting Shen and Wenyuan Yu and Xianzhong Shi and Xiaoming Huang and Xin Xu and Yan Kou and Yangyu Lv and Yifei Li and Yijing Liu and Yiming Wang and Yingya Zhang and Yitong Huang and Yong Li and You Wu and Yu Liu and Yulin Pan and Yun Zheng and Yuntao Hong and Yupeng Shi and Yutong Feng and Zeyinzi Jiang and Zhen Han and Zhi-Fan Wu and Ziyu Liu},
      year={2025},
      eprint={2503.20314},
      archivePrefix={arXiv},
      primaryClass={cs.CV},
      url={https://arxiv.org/abs/2503.20314}, 
}

@article{mechanistic_2026,
  title={A Mechanistic View on Video Generation as World Models: State and Dynamics},
  author={Wang, Luozhou and Chen, Zhifei and Du, Yihua and Yan, Dongyu and Ge, Wenhang and Shen, Guibao and Xu, Xinli and Wu, Leyi and Chen, Man and Xu, Tianshuo and others},
  journal={arXiv preprint arXiv:2601.17067},
  year={2026}
}

@misc{newton_2026,
      title={NEWTON: Agentic Planning for Physically Grounded Video Generation}, 
      author={Yuxiang Feng and Juncheng Wang and Chao Xu and Yijie Qian and Huihan Wang and Wenlong Hou and Yang Liu and Baigui Sun and Yong Liu and Shujun Wang},
      year={2026},
      eprint={2605.18396},
      archivePrefix={arXiv},
      primaryClass={cs.CV},
      url={https://arxiv.org/abs/2605.18396}, 
}

@misc{physics_cognition_2025,
      title={Exploring the Evolution of Physics Cognition in Video Generation: A Survey}, 
      author={Minghui Lin and Xiang Wang and Yishan Wang and Shu Wang and Fengqi Dai and Pengxiang Ding and Cunxiang Wang and Zhengrong Zuo and Nong Sang and Siteng Huang and Donglin Wang},
      year={2025},
      eprint={2503.21765},
      archivePrefix={arXiv},
      primaryClass={cs.CV},
      url={https://arxiv.org/abs/2503.21765}, 
}

@misc{unitemp_2026,
      title={UniTemp: Unlocking Video Generation in Any Temporal Order via Bidirectional Distillation}, 
      author={Lin Zhang and Sicheng Mo and Zefan Cai and Jinhong Lin and Zihao Lin and Jiuxiang Gu and Krishna Kumar Singh and Yuheng Li and Yin Li},
      year={2026},
      eprint={2606.18702},
      archivePrefix={arXiv},
      primaryClass={cs.CV},
      url={https://arxiv.org/abs/2606.18702}, 
}

@inproceedings{vace_2025,
    title = {VACE: All-in-One Video Creation and Editing},
    author = {Jiang, Zeyinzi and Han, Zhen and Mao, Chaojie and Zhang, Jingfeng and Pan, Yulin and Liu, Yu},
    booktitle = {Proceedings of the IEEE/CVF International Conference on Computer Vision},
    pages = {17191-17202},
    year = {2025}
}

@misc{ltxvideo_2024,
      title={LTX-Video: Realtime Video Latent Diffusion}, 
      author={Yoav HaCohen and Nisan Chiprut and Benny Brazowski and Daniel Shalem and Dudu Moshe and Eitan Richardson and Eran Levin and Guy Shiran and Nir Zabari and Ori Gordon and Poriya Panet and Sapir Weissbuch and Victor Kulikov and Yaki Bitterman and Zeev Melumian and Ofir Bibi},
      year={2024},
      eprint={2501.00103},
      archivePrefix={arXiv},
      primaryClass={cs.CV},
      url={https://arxiv.org/abs/2501.00103}, 
}

@InProceedings{zhu_2024_framewise,
  title={Generative Inbetweening through Frame-wise Conditions-Driven Video Generation},
  author={Zhu, Tianyi and Ren, Dongwei and Wang, Qilong and Wu, Xiaohe and Zuo, Wangmeng},
  booktitle = {Proceedings of the IEEE/CVF Conference on Computer Vision and Pattern Recognition (CVPR)},
  pages={27968--27978},
  year={2025}
}

@misc{videophy2_2025,
      title={VideoPhy-2: A Challenging Action-Centric Physical Commonsense Evaluation in Video Generation}, 
      author={Hritik Bansal and Clark Peng and Yonatan Bitton and Roman Goldenberg and Aditya Grover and Kai-Wei Chang},
      year={2025},
      eprint={2503.06800},
      archivePrefix={arXiv},
      primaryClass={cs.CV},
      url={https://arxiv.org/abs/2503.06800}, 
}

@InProceedings{cect_2026,
  title={Chain of Event-Centric Causal Thought for Physically Plausible Video Generation},
  author={Wang, Zixuan and Hu, Yixin and Wang, Haolan and Chen, Feng and Liu, Yan and Li, Wen and Lei, Yinjie},
  booktitle={Proceedings of the IEEE/CVF Conference on Computer Vision and Pattern Recognition (CVPR)},
  eprint={2603.09094},
  archivePrefix={arXiv},
  year={2026}
}

@InProceedings{phygenbench_2024,
      title={Towards World Simulator: Crafting Physical Commonsense-Based Benchmark for Video Generation}, 
      author={Fanqing Meng and Jiaqi Liao and Xinyu Tan and Wenqi Shao and Quanfeng Lu and Kaipeng Zhang and Yu Cheng and Dianqi Li and Yu Qiao and Ping Luo},
      year={2025},
      booktitle={Proceedings of the 42nd International Conference on Machine Learning (ICML)},
      series={Proceedings of Machine Learning Research},
      volume={267},
      pages={43781--43806},
      publisher={PMLR},
}

@inproceedings{physgen_2024,
      title={PhysGen: Rigid-Body Physics-Grounded Image-to-Video Generation},
      author={Liu, Shaowei and Ren, Zhongzheng and Gupta, Saurabh and Wang, Shenlong},
      booktitle={European Conference on Computer Vision (ECCV)},
      year={2024}
    }

@misc{physagent_2026,
      title={PhysAgent: Reflective Agentic Physics Control for Physically Plausible Video Generation}, 
      author={Qirui Li and Jinkun Hao and Yibo Li and Ran Yi and Paul L. Rosin and Yu-Kun Lai},
      year={2026},
      eprint={2607.16355},
      archivePrefix={arXiv},
      primaryClass={cs.CV},
      url={https://arxiv.org/abs/2607.16355}, 
}

@inproceedings{wang_2024_inbetweening,
  title={Generative Inbetweening: Adapting Image-to-Video Models for Keyframe Interpolation},
  author={Wang, Xiaojuan and Zhou, Boyang and Curless, Brian and Kemelmacher-Shlizerman, Ira and Holynski, Aleksander and Seitz, Steven M},
  booktitle={International Conference on Learning Representations (ICLR)},
  year={2025}
}

@misc{living_worlds_2026,
      title={Authoring for Living Worlds: Tool-Constrained LLM Agents for Executable Multi-Actor Scenarios}, 
      author={Nicolae Cudlenco and Mihai Masala and Marius Leordeanu},
      year={2026},
      eprint={2604.10383},
      archivePrefix={arXiv},
      primaryClass={cs.CV},
      url={https://arxiv.org/abs/2604.10383}, 
}

@inproceedings{vdaworld_2026,
  title     = {VDAWorld: World Modelling via VLM-Directed Abstraction and Simulation},
  author    = {O'Mahony, Felix and Cipolla, Roberto and Tewari, Ayush},
  booktitle = {Proceedings of SIGGRAPH Asia},
  year      = {2026}
}

@misc{physicsiq_leaderboard_2026,
  author       = {{Anates Labs}},
  title        = {{Physics-IQ Verified leaderboard}},
  year         = {2026},
  howpublished = {\url{https://physics-iq-verified.anates.ai}},
  note         = {Image-to-video track, single generation, normalized cost; accessed 8 October 2026},
}

@misc{physicsiq_verified_2026,
      title={Physics-IQ Verified}, 
      author={Tim Rädsch and Yuki M Asano and Hilde Kuehne and Stefan Bauer and Priyank Jaini and Robert Geirhos and Carsten T. Lüth},
      year={2026},
      eprint={2606.18943},
      archivePrefix={arXiv},
      primaryClass={cs.CV},
      url={https://arxiv.org/abs/2606.18943}, 
}

@InProceedings{genie_2024,
  title = 	 {Genie: Generative Interactive Environments},
  author =       {Bruce, Jake and Dennis, Michael D and Edwards, Ashley and Parker-Holder, Jack and Shi, Yuge and Hughes, Edward and Lai, Matthew and Mavalankar, Aditi and Steigerwald, Richie and Apps, Chris and Aytar, Yusuf and Bechtle, Sarah Maria Elisabeth and Behbahani, Feryal and Chan, Stephanie C.Y. and Heess, Nicolas and Gonzalez, Lucy and Osindero, Simon and Ozair, Sherjil and Reed, Scott and Zhang, Jingwei and Zolna, Konrad and Clune, Jeff and Freitas, Nando De and Singh, Satinder and Rockt\"{a}schel, Tim},
  booktitle = 	 {Proceedings of the 41st International Conference on Machine Learning},
  pages = 	 {4603--4623},
  year = 	 {2024},
  editor = 	 {Salakhutdinov, Ruslan and Kolter, Zico and Heller, Katherine and Weller, Adrian and Oliver, Nuria and Scarlett, Jonathan and Berkenkamp, Felix},
  volume = 	 {235},
  series = 	 {Proceedings of Machine Learning Research},
  month = 	 {21--27 Jul},
  publisher =    {PMLR},
  url = 	 {https://proceedings.mlr.press/v235/bruce24a.html}
}

@misc{qwen3_vl,
      title={Qwen3-VL Technical Report}, 
      author={Shuai Bai and Yuxuan Cai and Ruizhe Chen and Keqin Chen and Xionghui Chen and Zesen Cheng and Lianghao Deng and Wei Ding and Chang Gao and Chunjiang Ge and Wenbin Ge and Zhifang Guo and Qidong Huang and Jie Huang and Fei Huang and Binyuan Hui and Shutong Jiang and Zhaohai Li and Mingsheng Li and Mei Li and Kaixin Li and Zicheng Lin and Junyang Lin and Xuejing Liu and Jiawei Liu and Chenglong Liu and Yang Liu and Dayiheng Liu and Shixuan Liu and Dunjie Lu and Ruilin Luo and Chenxu Lv and Rui Men and Lingchen Meng and Xuancheng Ren and Xingzhang Ren and Sibo Song and Yuchong Sun and Jun Tang and Jianhong Tu and Jianqiang Wan and Peng Wang and Pengfei Wang and Qiuyue Wang and Yuxuan Wang and Tianbao Xie and Yiheng Xu and Haiyang Xu and Jin Xu and Zhibo Yang and Mingkun Yang and Jianxin Yang and An Yang and Bowen Yu and Fei Zhang and Hang Zhang and Xi Zhang and Bo Zheng and Humen Zhong and Jingren Zhou and Fan Zhou and Jing Zhou and Yuanzhi Zhu and Ke Zhu},
      year={2025},
      eprint={2511.21631},
      archivePrefix={arXiv},
      primaryClass={cs.CV},
      url={https://arxiv.org/abs/2511.21631}, 
}

@misc{qwen_image21,
  author       = {{Qwen Team}},
  title        = {{Qwen-Image-2.1}},
  year         = {2026},
  howpublished = {\url{https://qwen.ai/blog?id=qwen-image-2.1}},
}

@inproceedings{phyworldbench_2025,
  title={PhyWorldBench: A Comprehensive Evaluation of Physical Realism in Text-to-Video Models},
  author={Gu, Jing and Liu, Xian and Zeng, Yu and Nagarajan, Ashwin and Zhu, Fangrui and Hong, Daniel and Fan, Yue and Yan, Qianqi and Zhou, Kaiwen and Liu, Ming-Yu and Wang, Xin Eric},
  booktitle={International Conference on Learning Representations (ICLR)},
  year={2026}
}

@misc{sora_2024,
  author       = {{OpenAI}},
  title        = {{Video generation models as world simulators}},
  year         = {2024},
  howpublished = {\url{https://openai.com/index/video-generation-models-as-world-simulators/}},
}

@misc{cosmos3_2026,
      title={Cosmos 3: Omnimodal World Models for Physical AI}, 
      author={{NVIDIA} and Aditi and Niket Agarwal and Arslan Ali and Jon Allen and Martin Antolini and Adeline Aubame and Alisson Azzolini and Junjie Bai and Maciej Bala and Yogesh Balaji and Josh Bapst and Aarti Basant and Mukesh Beladiya and Mohammad Qazim Bhat and Zaid Pervaiz Bhat and Dan Blick and Vanni Brighella and Han Cai and Tiffany Cai and Eric Cameracci and Jiaxin Cao and Yulong Cao and Mark Carlson and Carlos Casanova and Ting-Yun Chang and Yan Chang and Yu-Wei Chao and Prithvijit Chattopadhyay and Roshan Chaudhari and Chieh-Yun Chen and Junyu Chen and Ke Chen and Qizhi Chen and Wenkai Chen and Xiaotong Chen and Yu Chen and An-Chieh Cheng and Click Cheng and Xiu Chia and Jeana Choi and Chaeyeon Chung and Wenyan Cong and Yin Cui and Magdalena Dadela and Nalin Dadhich and Wenliang Dai and Joyjit Daw and Alperen Degirmenci and Rodrigo Vieira Del Monte and Robert Denomme and Sameer Dharur and Marco Di Lucca and Ke Ding and Wenhao Ding and Yifan Ding and Yuzhu Dong and Nicole Drumheller and Yilun Du and Aigul Dzhumamuratova and Aleksandr Efitorov and Hamid Eghbalzadeh and Naomi Eigbe and Imad El Hanafi and Hassan Eslami and Benedikt Falk and Jiaojiao Fan and Jim Fan and Amol Fasale and Sergiy Fefilatyev and Liang Feng and Francesco Ferroni and Sanja Fidler and Xiao Fu and Vikram Fugro and Prashant Gaikwad and TJ Galda and Katelyn Gao and Yihuai Gao and Wenhang Ge and Sreyan Ghosh and Arushi Goel and Vivek Goel and Akash Gokul and Rama Govindaraju and Jinwei Gu and Miguel Guerrero and Elfie Guo and Aryaman Gupta and Siddharth Gururani and Hugo Hadfield and Song Han and Ankur Handa and Zekun Hao and Mohammad Harrim and Ali Hassani and Nathan Hayes-Roth and Yufan He and Chris Helvig and Cyrus Hogg and Madison Huang and Michael Huang and Sophia Huang and Yufan Huang and Jacob Huffman and DeLesley Hutchins and Suneel Indupuru and Boris Ivanovic and Arihant Jain and Joel Jang and Ryan Ji and Yanan Jian and Dongfu Jiang and Jingyi Jin and Atharva Joshi and Nikhilesh Joshi and Pranjali Joshi and Andy Ju and Jaehun Jung and Weiwei Kang and Scott Kassekert and Jan Kautz and Ashna Khetan and Julia Kiczka and Slawek Kierat and Gwanghyun Kim and Kuno Kim and Sunny Kim and Kezhi Kong and Xin Kong and Zhifeng Kong and Tomasz Kornuta and Egor Krivov and Hui Kuang and Saurav Kumar and Chia-Wen Kuo and George Kurian and Wojciech Kutak and JF Lafleche and Himangshu Lahkar and Omar Laymoun and Jayjun Lee and Sanggil Lee and Gabriele Leone and Boyi Li and Freya Li and Jiajun Li and Jinfeng Li and Ling Li and Pengcheng Li and Shangru Li and Tingle Li and Xiaolong Li and Xuan Li and Zhaoshuo Li and Zhiqi Li and Hao Liang and Maosheng Liao and Chen-Hsuan Lin and Tsung-Yi Lin and Ming-Yu Liu and Sifei Liu and Zihan Liu and Hai Loc Lu and Xiangyu Lu and Alice Luo and Ruipu Luo and Wenjie Luo and Jiangran Lyu and Martin Ding Ma and Nic Ma and Qianli Ma and Dawid Majchrowski and Louis Marcoux and Miguel Martin and Qing Miao and Ashkan Mirzaei and Shreyas Misra and Kaichun Mo and Durra Mohsin and Hyejin Moon and Pawel Morkisz and Saeid Motiian and Kirill Motkov and Seungjun Nah and Yashraj Narang and Deepak Narayanan and Thabang Ngazimbi and Julian Ouyang and Shubham Pachori and David Page and Yatian Pang and Sehwi Park and Mahesh Patekar and Mostofa Patwary and Marco Pavone and Trung Pham and Wei Ping and Soha Pouya and Shrimai Prabhumoye and Varun Praveen and Delin Qu and Hesam Rabeti and Morteza Ramezanali and Marilyn Reeb and Xuanchi Ren and Kristen Rumley and Wojciech Rymer and Jun Saito and Yeongho Seol and John Shao and Piyush Shekdar and Tianwei Shen and Humphrey Shi and Min Shi and Stella Shi and Kevin Shih and Mohammad Shoeybi and Mateusz Sieniawski and Shuran Song and Alexander Sotelo and Amir Sotoodeh and Sunil Srinivasa and Vignesh Srinivasakumar and Bartosz Stefaniak and Rahul Heinrich Steiger and Shangkun Sun and Jiaxiang Tang and Shitao Tang and Yangyang Tang and Yue Tang and Tolou Tavakkoli and Kayley Ting and Krzysztof Tomala and Wei-Cheng Tseng and Jibin Varghese and Sergei Vasilev and Thomas Volk and Raju Wagwani and Roger Waleffe and Andrew Z. Wang and Boxiang Wang and Haoxiang Wang and Qiao Wang and Shihao Wang and Shijie Wang and Ting-Chun Wang and Yan Wang and Yu Wang and Rohit Watve and David Wehr and Fangyin Wei and Xinshuo Weng and Jay Zhangjie Wu and Kedi Wu and Hongchi Xia and Summer Xiao and Tianjun Xiao and Kevin Xie and Daguang Xu and Jiashu Xu and Mengyao Xu and Ruqing Xu and Xingqian Xu and Yao Xu and Dinghao Yang and Dong Yang and Hans Yang and Xiaodong Yang and Xuning Yang and Yichu Yang and Yurong You and Zhiding Yu and Hao Yuan and Simon Yuen and Xiaohui Zeng and Pengcuo Zeren and Cindy Zha and Haotian Zhang and Jenny Zhang and Jing Zhang and Liangkai Zhang and Paris Zhang and Shun Zhang and Xuanmeng Zhang and Zhizheng Zhang and Ann Zhao and Yilin Zhao and Yuliya Zhautouskaya and Charles Zhou and Fengzhe Zhou and Shilin Zhu and Yuke Zhu and Dima Zhylko and Artur Zolkowski},
      year={2026},
      eprint={2606.02800},
      archivePrefix={arXiv},
      primaryClass={cs.CV},
      url={https://arxiv.org/abs/2606.02800}, 
}

\clearpage
\appendix

\section{Further Results}

\begin{figure*}[p]
    \centering
    \includegraphics[width=\linewidth,height=0.88\textheight,keepaspectratio]{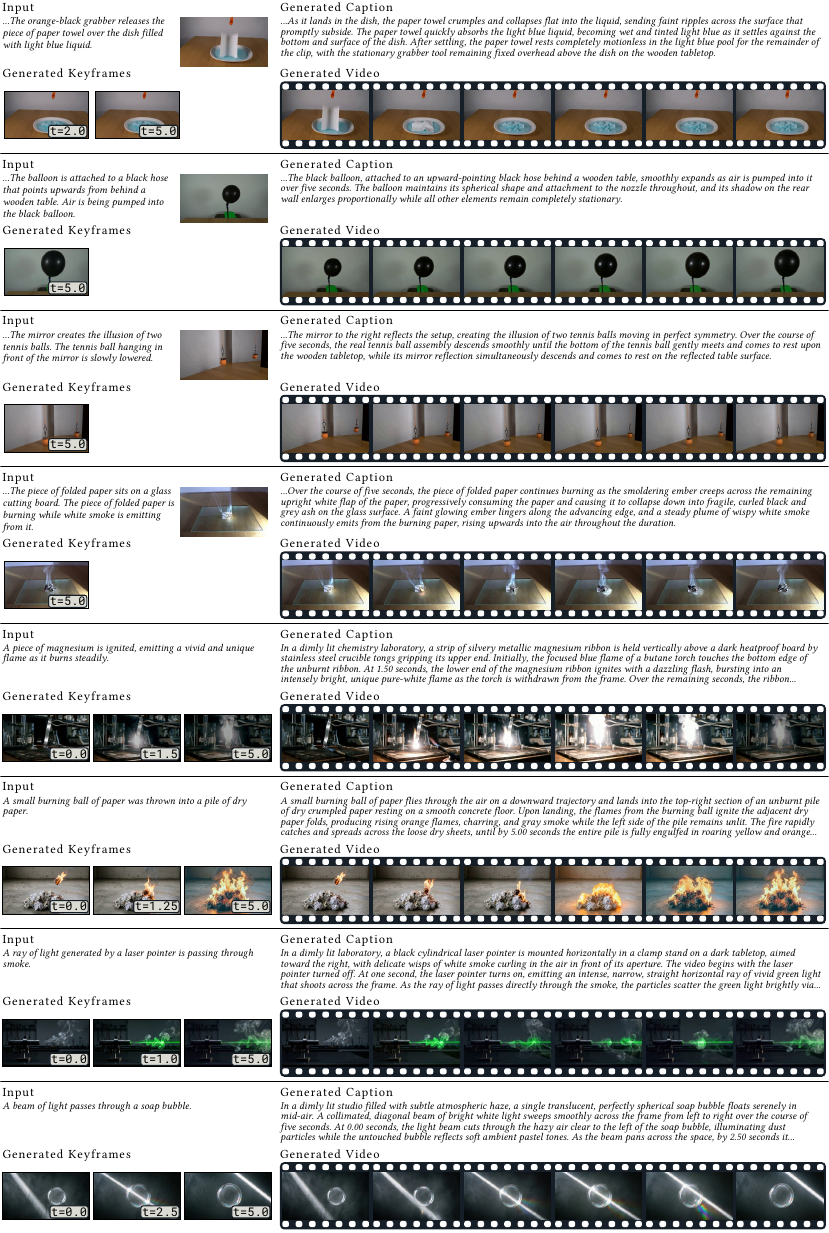}
    \caption{%
        Further results. Top four: Physics-IQ Verified (centre view, run 1), where the input is the benchmark's conditioning frame and description, and keyframes are placed where each scene's physics requires. Bottom four: PhyGenBench, where the input is the text prompt alone and the opening frame at $t=0$ is generated by our agent.
    }
    \label{fig:appendix_results}
\end{figure*}

\begin{figure*}[tb]
    \centering
    \includegraphics[width=\linewidth,trim={0 0 0 368.46pt},clip]{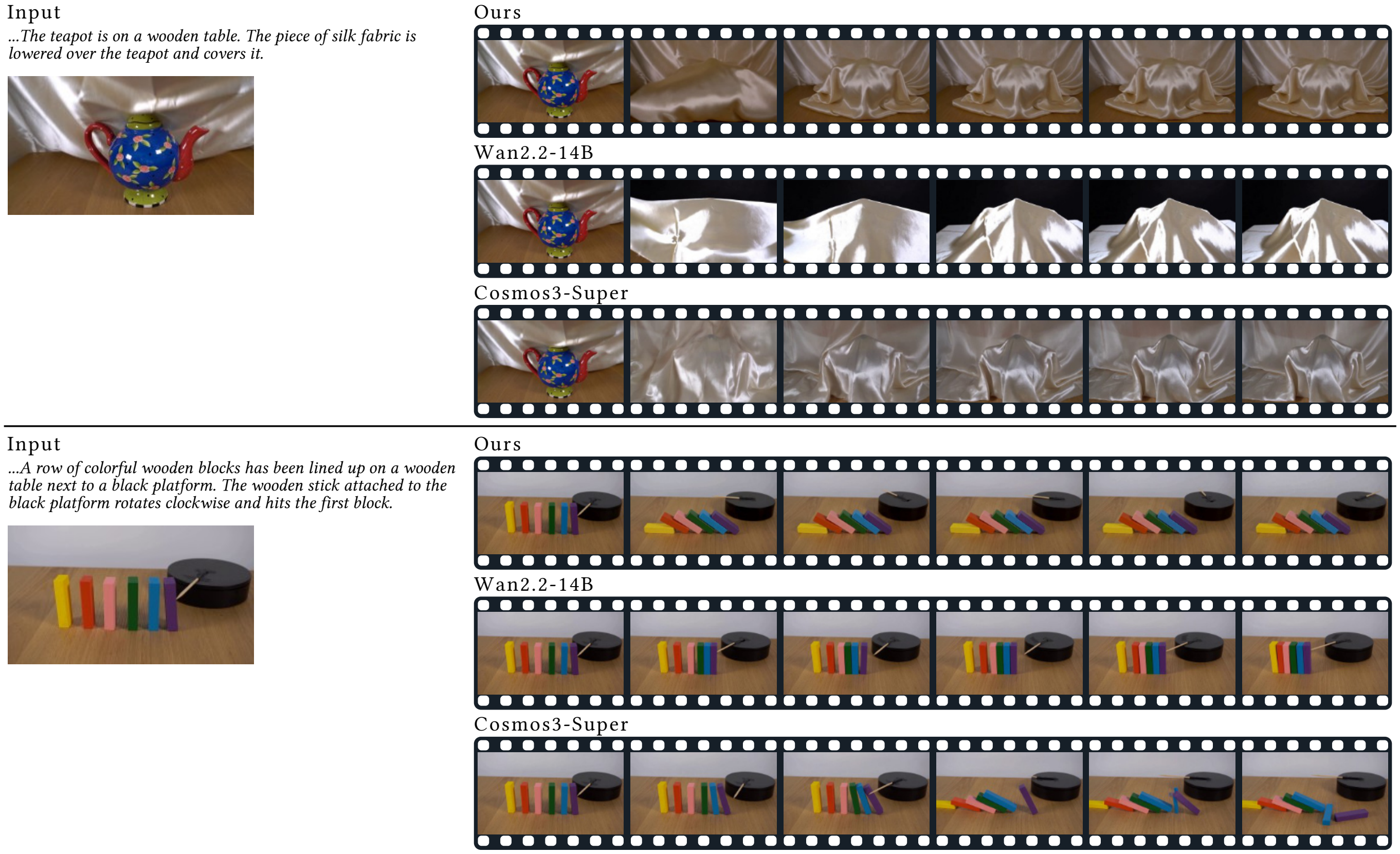}
    \caption{%
        Further qualitative comparisons on Physics-IQ Verified (centre view), continuing Figure~\ref{fig:qualitative_comparison}, against Wan2.2-14B and Cosmos3-Super from the same input frame.
    }
    \label{fig:appendix_comparison}
\end{figure*}

We show further results in Figure~\ref{fig:appendix_results}, and further comparisons on Physics-IQ Verified against unmodified Wan2.2-14B and Cosmos3-Super in Figure~\ref{fig:appendix_comparison}, continuing Figure~\ref{fig:qualitative_comparison} of the main paper. Every method is conditioned on the same input.
These highlight that our approach works in a diverse range of settings, including those requiring complex reasoning such as the Tyndall effect, the dispersion of light by a soap film, the colour of a magnesium flame and the spread of combustion.
All of this is facilitated through reasoning with a language model, and the successful integration of that model with a video model.

\section{Cost per Video}
\label{sec:cost}

Table~\ref{tab:cost} breaks down the cost per video quoted alongside Table~\ref{tab:phygenbench}.
Token counts for \methodname{} are measured over complete benchmark runs: the 66 centre-view scenes of Physics-IQ Verified and the 160 prompts of PhyGenBench.
For NEWTON\footref{fn:opennewton} we ran OpenNEWTON with its default settings on eight PhyGenBench prompts, accessing Seedance 2.0 Fast through BytePlus ModelArk; it uses an untrained GPT-5.5 planner rather than the trained planner of \citet{newton_2026}.

Prices are the standard-tier list prices on 8 October 2026.
Gemini 3.8 Flash~\citep{deepmind_2026} costs \$0.75 per million input tokens and \$3.75 per million output tokens, with thinking tokens billed as output.
Gemini 3.1 Flash Image~\citep{deepmind_2026b} costs \$0.50 per million input tokens and \$60 per million image-output tokens, about \$0.067 per image at the resolutions used here.
GPT-5.5 costs \$5 and \$30 per million input and output tokens.
Seedance 2.0 Fast bills a 5-second 720p video as 108,900 tokens at \$5.60 per million, or 195,300 tokens at \$3.30 per million when a reference video is supplied, so each video costs about \$0.61.
For Wan2.2-14B we use the normalized per-video cost listed on the Physics-IQ Verified leaderboard~\citep{physicsiq_leaderboard_2026}, which scales its reported \$0.11 at 16\,fps to 24\,fps; keyframe conditioning adds no denoising steps, so \methodname{}'s generation costs the same as unmodified Wan2.2-14B.
These are the per-video costs plotted in Figure~\ref{fig:cost_frontier}, which places \methodname{} against the models of Table~\ref{tab:leaderboard} on the leaderboard's cost--score chart.

The difference between the two methods is set mainly by the number of videos generated.
NEWTON generates a text-only baseline and then a further video for every correction cycle, between two and five per prompt in our runs, whereas \methodname{} generates exactly one.

\section{Agent Tools}
\label{sec:tools}

The planning agent of Section~\ref{sec:keyframe_module} acts entirely through the following tools.
At each step it chooses which to call, so the number of iterations spent on a scene is set by the agent rather than fixed in advance.

\begin{enumerate}
    \item \texttt{render\_\allowbreak edit(instruction)} calls the image edit model, passing it both a textual edit instruction and the latest keyframe.
    \item \texttt{view\_\allowbreak image()} passes the render back to the agent, allowing it to view the image.
    \item \texttt{evaluate\_\allowbreak render(evaluation)} requires the image model to evaluate the render before deciding whether to add it to the keyframe stack. This forces the model to consider whether the render adheres to its instructions. \texttt{add\_\allowbreak keyframe(time)} cannot be called until this method has first been called on a given render.
    \item \texttt{add\_\allowbreak keyframe(time)} adds the rendering to the list of accepted keyframes with a timestamp provided by the model. This keyframe then becomes the latest keyframe, which is edited by subsequent calls of \texttt{render\_\allowbreak edit(instruction)}.
    \item \texttt{finish(expanded\_\allowbreak caption)} ends the keyframe generation loop, passing the complete stack of keyframes and the expanded caption to the video model.

\end{enumerate}

\section{Planner prompt}
\label{sec:prompt}

The complete brief given to the planning agent for image-to-video generation. The
benchmark's scene description is appended in place of {\fontfamily{lmtt}\selectfont <scene description>}; all
other text is sent verbatim.

{\ttfamily\scriptsize\raggedright
\setlength{\parindent}{0pt}\setlength{\parskip}{0pt}
You are given one frame from a video, and a short description of the scene.

\vspace{5pt}

The frame is the **handover frame**: the video continues from this moment for 5.00 more seconds. It is not necessarily the start of the action. Footage precedes it, and motion may already be under way in the frame you are looking at. The description may be written as though nothing has happened yet.

\vspace{5pt}

Your job is to build a **stack of keyframes**: pictures of this scene at fixed times, which together say what happens over those 5.00 seconds. The stack already holds the handover frame at 0.00s.

\vspace{5pt}

You work through tools, and you can see what you make. Use that: look at the photograph before you commit it.

\vspace{5pt}

\#\# What you are working with

\vspace{5pt}

The **stack** is the keyframes you have committed. Every photograph you render is made by editing the picture on top of the stack, so the arrangement one moment settles into is the arrangement the next moment starts from.

\vspace{5pt}

\#\# Working out what happens

\vspace{5pt}

- Reason physically, from what the frame actually shows: the objects present, their sizes and likely masses, what is in contact with what, what is already in motion and in which direction.

- Reason about where things end up.

- If an interaction seems likely but will not in fact occur, do not depict it.

- Do not invent objects, cuts or camera movement unless the description says so.

\vspace{5pt}

\#\# Choosing the keyframes to add to the stack

\vspace{5pt}

A keyframe is a moment worth pinning: a boundary between distinct phases of motion. Something arrives somewhere and the situation changes qualitatively — an object lands, stops, breaks contact, reverses, tips past balance, or reaches a configuration it then holds.

\vspace{5pt}

Choose them sparsely. Most scenes need one or two. Use three or four only when the span genuinely contains that many distinct phases. Do not narrate the mechanism: where several things happen in quick succession as part of one continuous event, that is one keyframe, the state the event leaves behind.

\vspace{5pt}

**Always include a keyframe at 5.00s**, even if this moment does not represent the boundary of a distinct phase.

\vspace{5pt}

\#\# Rendering and committing

\vspace{5pt}

\textasciigrave{}render\_edit\textasciigrave{} takes a text instruction and the keyframe at the top of the stack, and asks an image model to edit the keyframe according to the instruction. You should specify the edit in terms of what should and should not change to accomplish the edit. You should consider that your instruction should direct the edit relative to the top keyframe, not the handover frame unless the top keyframe is not the handover frame and no renders have already been added to the stack.

\vspace{5pt}

Renders are the one expensive thing you do, and the budget is small: **8 of them for this scene**, and 24 turns in total. Every tool result ends with what is left of both.

\vspace{5pt}

**Every render must be evaluated before you can do anything else.** As soon as one comes back, \textasciigrave{}evaluate\_render\textasciigrave{} is the only tool available to you: say what the picture actually shows, item by item. Count what is countable. Only once you have answered can you render again, or add it to the stack.

\vspace{5pt}

**Look at what came back, and anticipate that it might be wrong.** It may quietly change things it was not asked about — the most common failures are objects appearing or disappearing, object counts changing, an object ending up in a different place or a different orientation from the one you drew. Judge it item by item:

\vspace{5pt}

- Is everything you expect present in the photograph, and nothing else?

- Is each thing that moved where you expected it, at the orientation described?

- Has anything changed that you did not expect?

\vspace{5pt}

Nothing checks this after you. Whatever you push is what is accepted as a keyframe, and a keyframe that says something untrue about the scene is worse than one you spent another render getting right.

\vspace{5pt}

If it does not match, you have three options to correct it. You can adjust the edit instruction given to \textasciigrave{}render\_edit\textasciigrave{} to more strictly enforce the edit, or you can simply run the render again, since the rendering model is stochastic.

\vspace{5pt}

You have a numbered list of **images**. Image 0 is the handover frame; every render you make is added to it as image 1, 2, and so on, and they all stay available for the rest of the run.

\vspace{5pt}

\textasciigrave{}add\_keyframe\textasciigrave{} puts one of those images on the stack at a time. It uses the most recent image unless you name another, and it uses nothing up:

\vspace{5pt}

- If a second attempt comes back worse than the first, add the earlier one by its number rather than paying for a third. \textasciigrave{}view\_image\textasciigrave{} shows you one again if it has scrolled out of sight.

- If nothing changes between one moment and a later one, add the \_same\_ image at both times rather than rendering the same picture twice. It costs no render and cannot introduce a difference nobody asked for.

- If nothing changes at all, add image 0 — the handover frame — at the end of the clip.

\vspace{5pt}

Choosing between pictures you already have costs nothing. Making another costs a render.

\vspace{5pt}

\#\# Finishing

\vspace{5pt}

Call \textasciigrave{}finish\textasciigrave{} with one paragraph of plain description of the action in the whole 5.00 seconds. The description you give should expand on the content contained in the scene description. It should contain all of the information in the original description, with a detailed account of what happens in the 5.00 seconds after the handover frame, and no more. Content should only be removed if it refers to something that occurs before the handover frame, and would therefore not be relevant for the 5.00 seconds after it.

\vspace{5pt}

Scene description:

\textless{}scene description\textgreater{}
\par}

\begin{table*}[t]
\centering
\caption{%
  Cost per video in US dollars at list prices on 8 October 2026. Counts are means per
  video. The NEWTON total excludes its video verifier, whose usage was not metered.
}
\label{tab:cost}
\begin{tabular}{llcc}
\toprule
Component & Model & Physics-IQ & PhyGenBench \\
\midrule
\multicolumn{4}{l}{\emph{\methodname{}}} \\
\quad Planning agent   & Gemini 3.8 Flash       & 6.1 calls, 0.161  & 10.1 calls, 0.165 \\
\quad Keyframe edits   & Gemini 3.1 Flash Image & 1.7 images, 0.143 & 2.9 images, 0.258 \\
\quad Video generation & Wan2.2-14B             & 1 video, 0.165    & 1 video, 0.165    \\
\quad \textbf{Total}   &                        & \textbf{0.47}     & \textbf{0.59}     \\
\midrule
\multicolumn{4}{l}{NEWTON\footref{fn:opennewton}} \\
\quad Planning agent   & GPT-5.5                & ---               & 6.6 calls, 0.32   \\
\quad Keyframe images  & Gemini 3.1 Flash Image & ---               & 2.4 images, 0.16  \\
\quad Video generation & Seedance 2.0 Fast      & ---               & 4.0 videos, 2.45  \\
\quad Verifier         & Gemini 3.5 Flash       & ---               & not metered       \\
\quad \textbf{Total}   &                        & ---               & \textbf{2.93} \\
\bottomrule
\end{tabular}
\end{table*}

\end{document}